\documentclass[runningheads]{llncs}
\usepackage{graphicx}
\usepackage{subfig}
\usepackage{multirow}
\usepackage[dvipsnames]{xcolor}
\begin{document}
\title{Explaining AI-based Decision Support Systems using Concept Localization Maps}
% \title{Explainable Decision Support Systems using Concept Localization Maps\thanks{Supported by organization x.}}
\author{Adriano Lucieri\inst{1,2}\orcidID{0000-0003-1473-4745} \and
Muhammad Naseer Bajwa\inst{1,2}\orcidID{0000-0002-4821-1056} \and
Andreas Dengel\inst{1,2}\orcidID{0000-0002-6100-8255} \and
Sheraz Ahmed\inst{2}\orcidID{0000-0002-4239-6520}}
\authorrunning{A. Lucieri et al.}
\titlerunning{Explaining AI-based DSS using Concept Localization Maps}
\institute{TU Kaiserslautern, 67663 Kaiserslautern, Germany \and
German Research Center for Artificial Intelligence GmbH (DFKI), Trippstadter Straße 122, 67663 Kaiserslautern, Germany\\
\email{\{adriano.lucieri, naseer.bajwa, andreas.dengel, sheraz.ahmed\}@dfki.de}\\
%\url{http://www.springer.com/gp/computer-science/lncs}
}
\maketitle              % typeset the header of the contribution
\begin{abstract}
Human-centric explainability of AI-based Decision Support Systems (DSS) using visual input modalities is directly related to reliability and practicality of such algorithms. An otherwise accurate and robust DSS might not enjoy trust of experts in critical application areas if it is not able to provide reasonable justification of its predictions. This paper introduces Concept Localization Maps (CLMs), which is a novel approach towards explainable image classifiers employed as DSS. CLMs extend Concept Activation Vectors (CAVs) by locating significant regions corresponding to a learned concept in the latent space of a trained image classifier. They provide qualitative and quantitative assurance of a classifier's ability to learn and focus on similar concepts important for humans during image recognition. To better understand the effectiveness of the proposed method, we generated a new synthetic dataset called Simple Concept DataBase (SCDB) that includes annotations for 10 distinguishable concepts, and made it publicly available. We evaluated our proposed method on SCDB as well as a real-world dataset called CelebA. We achieved localization recall of above 80\% for most relevant concepts and average recall above 60\% for all concepts using \textit{SE-ResNeXt-50} on SCDB. Our results on both datasets show great promise of CLMs for easing acceptance of DSS in practice.

\end{abstract}

\section{Introduction}
Inherently inquisitive human nature prompts us to unfold and understand the rationale behind decisions taken by Artificial Intelligence (AI) based algorithms. This curiosity has led to the rise of Explainable Artificial Intelligence (XAI), which deals with making AI-based models considerably transparent and building trust on their predictions. Over the past few years, AI researchers are increasingly drawing their attention to this rapidly developing area of research not only because it is driven by human nature but also because legislations across the world are mandating explainability of AI-based solutions \cite{guo2019explainable,waltl2018explainable}.

The applications of XAI are at least as widespread as AI itself including medical image analysis for disease predictions \cite{Lucieri2020Interpretability}, text analytics \cite{qureshi2019eve}, industrial manufacturing \cite{rehse2019towards}, autonomous driving \cite{glomsrud2020trustworthy}, and insurance sector \cite{bloomberg2018}. Many of these application areas utilize visual inputs in the form of images or videos. Humans recognize images and videos by identifying and localizing various concepts that are associated with objects – for example concepts of shape (bananas are long and apples are round) and colour (bananas are generally yellow and apples are red or green). XAI methods dealing with images also employ a similar approach of identifying and localizing regions in the input space of a given image, which correspond strongly with presence or absence of a certain object, or concept associated with the object.

%Most of the solutions addressing explainability of AI-based algorithms like Deep Neural Networks (DNNs) are developed using simple datasets and under ideal conditions before testing them on challenging real-word datasets under realistic conditions. 
In this work we attempt to provide a new perspective on explainable image classifiers by introducing Concept Localization Maps (CLMs), which are generated to locate human-understandable concepts as learnt and encoded by a classifier in its latent space and map them to input space for a given image. These CLMs validate that the AI-based algorithm learned to focus on pertinent regions in the image while looking for relevant concepts. This work builds on Concept Activation Vectors (CAVs) proposed in~\cite{kim2017interpretability} and extends concept-based explanation methods by introducing concept localization. The contributions of this work are as follows:
\begin{itemize}
    \item We propose \textit{Concept Localization Maps} (CLMs) as a means to localize concepts learned by Deep Neural Networks (DNNs) based image classifiers.
    %human-understandable concepts \colorbox{yellow}{learnt from mappings to the DNN's latent space} \colorbox{yellow}{and to check a mappings validity.} \textcolor{red}{Rephrase}
    \item We develop a new synthetically generated dataset of geometric shapes with annotations for 10 concepts and segmentation maps called SCDB. This dataset mimics complex relationships between concepts and classes in real-world medical image analysis tasks and can assist research in classification and localization of concepts.
    %This dataset \textcolor{red}{[give it a name]} is inspired by complex challenges of dermatoscopic image classification tasks.
    \item We evaluate CLMs qualitatively and quantitatively using three different model architectures (\textit{VGG16}~\cite{Simonyan15}, \textit{ResNet50}~\cite{He2015}, and \textit{SE-ResNeXt-50}~\cite{hu2018squeeze}) trained on SCDB dataset to show that the proposed method works across different network architectures. We also demonstrate practicality of this method in real work applications by applying it on \textit{SE-ResNeXt-50} trained on CelebA dataset. %use CLMs for the explanation of a classifier for CelebA face dataset to demonstrate its practicality in real-world applications.
\end{itemize}

%\textcolor{red}{[Can be added if space permits]} The rest of the paper is organized as follows. Section~\ref{sec:RelatedWorks} gives an overview about related works in concept-based explanation and feature relevance maps. The proposed dataset and other datasets used in this work are introduced in section~\ref{sec:Datasets}. The method and variants of CLM are introduced in section~\ref{sec:ProposedApproach}. A qualitative and quantitative evaluation of CLM is provided in section~\ref{sec:Experiments} along with the investigation of a classifier for natural image classification. Section~\ref{sec:Conclusion} summarizes the contributions and suggests further directions for improvement and application of the work.

\section{Related Works}
\label{sec:RelatedWorks}

\subsection{Concept-based Explanation Methods}
Concept-based explanation methods try to semantically link meaningful human-understandable concepts to internal states of a trained DNN. Kim et al.~\cite{kim2017interpretability} developed a line of work on Concept Activation Vectors (CAVs), which is a concept learning paradigm that maps human-defined concepts to a neural network's latent space. 
%Their TCAV score metric can be used to quantify a concept's influence to the prediction of a target class on a global scale by evaluating the sensitivity of many local samples in the concept direction. 
Their work has been used in various medical image analysis applications including histopathology~\cite{graziani2018regression}, ophthalmology~\cite{graziani2019improved}, dermatology~\cite{Lucieri2020Interpretability}, and has been extended by an unsupervised variant that does not require pre-definition of concepts~\cite{Ghorbani2019ACE}. %followed by applications on regression problems in histopathology and opthalmology~\cite{graziani2018regression,graziani2019improved}, on dermatology~\cite{Lucieri2020Interpretability} and by an unsupervised variant of the paradigm that does not require concepts' pre-definition~\cite{Ghorbani2019ACE}.
Other works have aimed at decomposing the latent space of networks~\cite{zhou2018interpretable} or mapping concepts to specific units in a network~\cite{bau2017network,bau2019gandissect}.
Recently, Goyal et al.~\cite{goyal2019explaining} have paired the notion of concepts with counterfactual explanations using conditional Variational AutoEncoders (VAE) to generate counterfactual images for specific concepts.

%Our work aims at continuing the work of~\cite{kim2017interpretability} and expand the area of concept-based explanation methods by introducing the field of concept localization. 

\subsection{Feature Importance Estimators}
Saliency or relevance maps are popular means for local interpretation of a trained DNN. Different methodologies exist to generate maps that indicate importance of input features for a given prediction. Gradient-based attribution methods use backpropagation to trace output gradients back to the origin in the input feature space. These methods generate heatmaps by backpropagating the gradient either once through the network~\cite{bach2015pixel,selvaraju2016grad,simonyan2014deep,sundararajan2017axiomatic} or by using ensembling techniques~\cite{adebayo2018sanity,smilkov2017smoothgrad}, which reduces noise considerably. On the other hand, perturbation-based attribution methods generate heatmaps by perturbing an input sample and observing changes in the model's prediction score. A distinction can be drawn between methods that perturb the inputs randomly ~\cite{Petsiuk2018rise}, systematically~\cite{zeiler2014visualizing} or based on optimization~\cite{fong2017interpretable,fong2019understanding}. 
%Additionally, approximation-based methods approximate complex models locally using easily explainable predictors. LIME~\cite{ribeiro2016should} is one such method that also generates local heatmaps.
%Apart from base estimation methods~\cite{simonyan2014deep,selvaraju2016grad,sundararajan2017axiomatic,bach2015pixel} that obtain their heatmaps by backpropagating the network once, ensembling techniques like SmoothGrad (SG)~\cite{smilkov2017smoothgrad} and its variants VarGrad~\cite{adebayo2018sanity} and SmoothGrad-Squared (SG-SQ) showed to improve the quality of resulting heatmaps by removing noise.

%\subsubsection{Perturbation-based} attribution methods generate heatmaps or relevance masks by perturbing an input sample and observing changes in the model's prediction score. A distinction can be drawn between methods that perturb the inputs randomly ~\cite{Petsiuk2018rise}, systematically~\cite{zeiler2014visualizing} or based on optimization~\cite{fong2017interpretable,fong2019understanding}.

%\paragraph{Approximation-based} methods approximate complex models locally using easily explainable predictors. LIME~\cite{ribeiro2016should} is one such method that also generates local heatmaps.

\section{Datasets}
\label{sec:Datasets}

\subsection{SCDB - Simple Concept DataBase} 

%Research in XAI usually faces the problem that classification tasks are either too simple or too complex for researchers to easily draw conclusions about a model's behaviour, often requiring specific domain knowledge \textcolor{red}{[Needs Reference]}. 
Attribution methods proved to work well in simpler detection tasks where entities are spatially easy to separate~\cite{jolly2018convolutional,ghorbani2020deep,selvaraju2016grad} but often fail to provide meaningful explanations in more complex and convoluted domains like dermatology, where concepts indicative for the predicted classes are spatially overlapping.
%~\cite{schulz2016kompendium}
Therefore, we developed and released Simple Concept DataBase (SCDB)\footnote{https://github.com/adriano-lucieri/SCDB}, a new synthetic dataset of complex composition, inspired by the challenges of skin lesion classification using dermatoscopic images. In SCDB, skin lesions are modelled as randomly placed large geometric shapes (base shapes) on black background. These base shapes are randomly rotated and have varying sizes and colours. The \textit{disease biomarkers} indicative of the ground truth labels are given as combinations of smaller geometric shapes within a larger base shape. \textit{Biomarkers} can appear in a variety of colours, shapes, locations and orientations as well. Semi-transparent fill colour allows \textit{biomarkers} to spatially overlap. The dataset has two defined classes, C1 and C2, indicated by different combinations of \textit{biomarkers}. C1 is indicated by joint presence of concepts $\textit{hexagon} \wedge \textit{star}$ or $\textit{ellipse} \wedge \textit{star}$ or $\textit{triangle} \wedge \textit{ellipse} \wedge \textit{starmarker}$. C2 is indicated by joint presence of concepts $\textit{pentagon} \wedge \textit{tripod}$ or $\textit{star} \wedge \textit{tripod}$ or $\textit{rectangle} \wedge \textit{star} \wedge \textit{starmarker}$. In addition to the described combinations, further \textit{biomarkers} are randomly generated within the base shape, without violating the classification rules. Two more \textit{biomarkers} (i.e. \textit{cross} and \textit{line}) are randomly generated on the base shape without any relation to target classes. Finally, random shapes are generated outside of the base shape for further distraction.
%The dataset has two defined classes, C1 and C2, indicated by combinations of \textit{biomarkers} given in Table~\ref{tab:ToyDataset}. In addition to the described combinations, further \textit{biomarkers} are randomly generated within the base shape, without violating the classification rules. Two more \textit{biomarkers} (i.e. \textit{cross} and \textit{line}) are randomly generated on the base geometry without any relation to target classes. Finally, random shapes are generated outside of the base geometry for further distraction. 

% \begin{table}[!t]
% \centering
% \caption{Mapping of shape combinations clearly indicating target classes.}
% \label{tab:ToyDataset}
% \begin{tabular}{c|c|c}
% Target Class                            & C1                       & C2                     \\ \hline
% \multirow{3}{*}{\shortstack{Biomarker\\ ~~Combinations~~}} & hexagon, star                 & pentagon, tripod            \\
%                                         & ~~triangle, ellipse, starmarker~~ & ~~rectangle, star, starmarker~~ \\
%                                         & ellipse, star                 & tripod, star               
% \end{tabular}
% \end{table}

\begin{figure}[h!]
\centering
 \includegraphics[width=.95\textwidth]{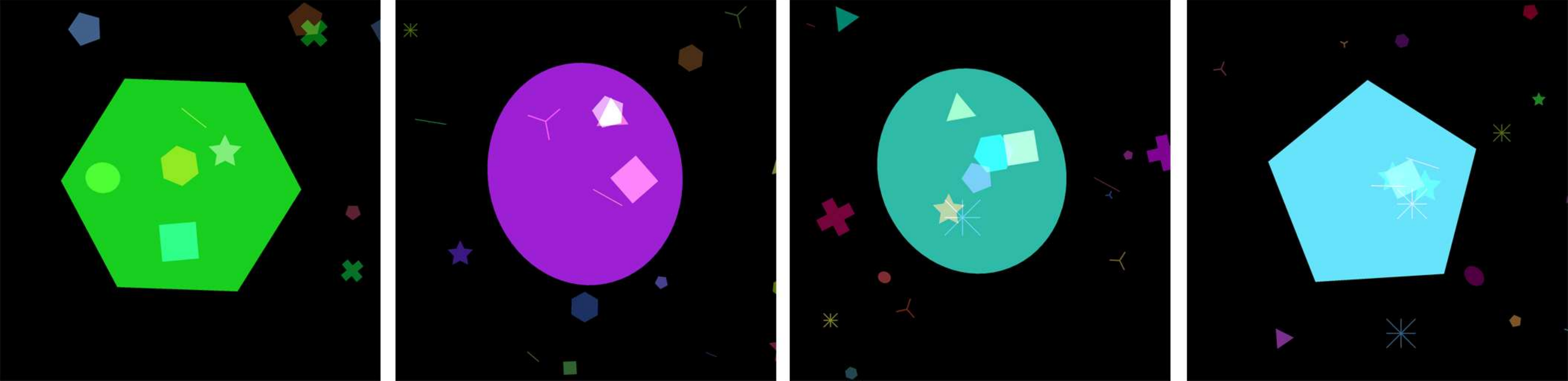}
\caption{Training samples from the proposed dataset.} \label{fig:ToyDataset}
\end{figure}

The dataset consists of 7\,500 samples for binary image classification and is divided into train, validation and test splits of 4\,800, 1\,200, 1\,500 samples respectively. Another 6\,000 images are provided separately for concept training. Along with each images, binary segmentation maps are provided for every concept present in the image in order to evaluate concept localization performance. Segmentation maps are provided as the smallest circular area enclosing the \textit{biomarker}. Fig.~\ref{fig:ToyDataset} shows examples of dataset samples.

%Figure~\ref{fig:ToyDataset} shows examples of dataset samples.
%along with their segmentation maps.

% The proposed dataset is a simplification of skin lesion datasets like PH2~\cite{mendoncca2013ph} and derm7pt~\cite{Kawahara2018-7pt}. Therefore, some more advanced relations like the effect of lesion shape, distribution,  as well as biomarker distribution, location and  on the target variables have been neglected for the time being. The dataset is released on github~\footnote{Link will be provided after acceptance.}.

% \begin{itemize}
%     \item Was war die Motivation? --> Ich hab einen Datensatz gebraucht der nicht zu basic ist (Objektdetektion) aber auch nicht so komplex dass nur experten ihn interpretieren können.
%     \item Deswegen haben wir in Anlehnung an das Problem der Dermatoskopie einen Datensatz entwickelt, der 1. verschiedene geometrische Formen als Biomarker enthält, 2. Die klassen sich durch erkennung von Kombinationen von geometrischen Mustern ergeben, 3. Die geometrien überlappen können (wie im echten fall), 4. Farben steuerbar sind, 5. Objekte nur in einer relevanten Zone für Diagnose entscheident sind, 6. Außerhalb zur Ablenkung formen vorkommen, die nicht relevant sind, 7. Innerhalb zur Ablenkung Formen vorkommen die nicht relevant sind.
%     \item Which rules it follows and which components exist
% \end{itemize}

\subsection{CelebA}
The proposed method can be evaluated on those real-world datasets that provide concept annotations. %Evaluation of the proposed method on real world datasets is not trivial as concept annotations are required.
%as well as segmentation maps for quantification. 
%However, many publicly available datasets that provide concept or part annotations are designed specifically for segmentation tasks such as ADE20K \cite{zhou2016semantic} and COCO~\cite{lin2014microsoft}. 
\textit{CelebA}~\cite{liu2015faceattributes} is such a dataset of faces with rich face attribute annotations. The dataset contains 202\,599 images of 10\,177 identities each annotated with regards to 40 binary attributes. The dataset has been split in train, validation, and test splits of 129\,664, 32\,415 and 40\,520 samples, evenly divided with respect to gender labels. 

An important aspect to consider while selecting datasets was to find a dataset that not only contains annotations of fine-grained concepts but also high-level concepts that can be indicated by solving an interim task of fine-grained concept detection. We chose CelebA because the gender annotation allows for solving a non-trivial classification task that relies on some of the remaining annotated concepts like \textit{baldness}, \textit{mustache}, \textit{lipstick} and \textit{makeup}, which statistically indicate the gender in the given data distribution.

% \subsubsection{Dermatology Datasets} Two skin lesion datasets with rich dermatoscopic criteria annotations, namely \textit{PH\textsuperscript{2}} dataset~\cite{mendoncca2013ph} and Seven-Point Checklist Dermatology dataset~\cite{Kawahara2018-7pt} (abbreviated as \textit{derm7pt}) are used. Both datasets are merged for evaluation, resulting in 1\,023 images of melanoma and nevus.

\section{CLM - The Proposed Approach}
\label{sec:ProposedApproach}

% Given an input image \( x \in X \) containing a given concept \( C \), a trained DNN \( f(x; \theta) \) with optimal weights \( \theta \) and the linear concept classifier of concept \( C \) \( g_C(f_l(x; \theta), \gamma) \) with optimal weights \( \gamma \), we want to obtain a localization map \( m \) that locates the essential region in the input image, responsible for the prediction score of the concept classifier. We tested a row of approaches as follows.

The CLM method obtains a localization map \( m_{Cl} \) for a concept \( C \) learnt on DNN's layer \( l \), that locates the relevant region essential for the prediction of a concept classifier \( g_C(f_l(x; \theta)) \) given an input image \( x \in X \). The linear concept classifier \( g_C \) generates a concept score for concept \( C \) given a latent vector of trained DNN \( f_l(x; \theta) \) with optimal weights \( \theta \) at layer \( l \). The resulting map \( m_{Cl} \) corresponds to the region in the latent space of DNN that encodes the concept~$C$.

\subsection{g-CLM}
\label{sec:ApproachGradient}
In order to be able to apply gradient-based attribution methods for concept localization we must find a binary mask \( m_{binC} \) that filters out latent dimensions, which contribute least to the classification of concept \( C \). For each concept we determine those dimensions by thresholding the concept classifier's weight vector \( v_C \), also known as CAV. High absolute weight values imply a strong influence of the latent feature dimension to the concept prediction and shall thus be retained. Therefore, a threshold value \( T_C \) is computed automatically based on 90\% percentile of weight values in \( v_C \). 
%We compare two binarization schemes for latent feature dimension masking.

%\paragraph{Independent Dimension Masking} In the first case, latent feature dimensions are masked independently based on the binary mask obtained through \( T_C \).

%\paragraph{Feature Map Wise Masking} In this case, only complete latent feature maps are retained or masked. This decision is based on ...

Gradient-based attribution methods are applied once the latent feature dimension is masked and the concept-relevant latent subset \( f_{lC}(x, \theta) \) is obtained. The methods evaluated in our work apply SmoothGrad\textsuperscript{2} (SG-SQ) and VarGrad as ensembling approaches using plain input gradients as base-estimator defined in equations~\ref{eq:SG} and~\ref{eq:VAR}. The noise vector \( g_j \sim \mathcal{N}(0,\,\sigma^{2})\) is drawn from a normal distribution and sampling is repeated \( N = 15 \) times. SG-SQ and VarGrad were proven to be superior to classical SG in~\cite{hooker2019benchmark} in terms of trustworthiness and spatial density of attribution maps. Henceforth, all experiments referring to gradient-based CLM will be denoted by g-CLM.

% \noindent\begin{minipage}{.5\linewidth}
% \begin{equation}
% \label{eq:SG}
%     m_{Cl} = \frac{1}{N} \sum_{j=1}^{N} \left( \frac{\mathop{d} f_lC(x_{i,j}+g_{j})}{\mathop{d} x_{i,j}+g_{j}} \right)^2
% \end{equation}
% \end{minipage}%
% \noindent\begin{minipage}{.5\linewidth}
% \begin{equation}
% \label{eq:VAR}
%     m_{Cl} = Var\left( \frac{\mathop{d} f_lC(x_{i,j}+g_{j})}{\mathop{d} x_{i,j}+g_{j}} \right)
% \end{equation}
% \end{minipage}
    
\begin{equation}
    \label{eq:SG}
    m_{Cl} = \frac{1}{N} \sum_{j=1}^{N} \left( \frac{\mathop{d} f_lC(x_{i,j}+g_{j})}{\mathop{d} x_{i,j}+g_{j}} \right)^2
\end{equation}
\begin{equation}
    \label{eq:VAR}
    m_{Cl} = Var\left( \frac{\mathop{d} f_lC(x_{i,j}+g_{j})}{\mathop{d} x_{i,j}+g_{j}} \right)
\end{equation}

% \begin{equation}
% Integrated Gradient, if I will use it.
% \end{equation}

% \begin{equation}
% m_C = \sum_{i=1}^{J} ( BE(x_i)^2 ) 
% \end{equation}

% \begin{equation}
% m_C = Var(BE(x_i)) 
% \end{equation}

\subsection{p-CLM}
The application of perturbation-based attribution methods requires local manipulation of the input image to observe changes in prediction output. In the case of CLM, the output is the predicted score of the concept classifier. The systematic occlusion method from~\cite{zeiler2014visualizing} 
%as well as the optimization-based method from~\cite{fong2019understanding} were 
is used in the experimentation section. For all reported experiments, a patch-size of \( 30 \) and stride of \( 10 \) is used, as it provides a good trade-off between smoothness of obtained maps and localization performance. Occluded areas are replaced by black patches. All experiments referring to the perturbation-based CLM method are denoted as p-CLM.

% \subsection{Concept Accordance Metric}
% In order to evaluate the performance of a given method, we use Intersection over Union (IoU) as concept accordance metric to evaluate the concept localization performance in case of available concept segmentation masks. The metric is 

\section{Experiments}
\label{sec:Experiments}
This section provides a quantitative and qualitative evaluation of CLM on SCDB and CelebA to prove its feasibility in practice.  

\subsection{Experimental Setup}
\label{sec:ExperimentSettings}
Three DNN types, namely VGG16, ResNet50 and SE-ResNeXt-50 are examined using CLM in order to study the influence of architectural complexity on concept representation and localization. All models were initialized with weights pre-trained on ImageNet~\cite{deng2009imagenet}. Hyperparameter tuning on optimizer and initial learning rate resulted in best results for optimization using RMSprop~\cite{Tieleman2012} with an initial learning rate of \( \mathrm{10}^{-4} \), learning rate decay of factor \( 0.5 \) with a tolerance of \( 5 \) epochs and early stopping of \( 10 \) epochs with a maximum number of \( 100 \) training epochs.

\begin{figure}[h!]
\centering
\subfloat[Random images along with their generated CLMs for concept \textit{ellipse}. Middle row shows g-CLM (SG-SQ) and lower row shows p-CLM.]{\label{fig:ToyEvalA} \includegraphics[width=.8\textwidth]{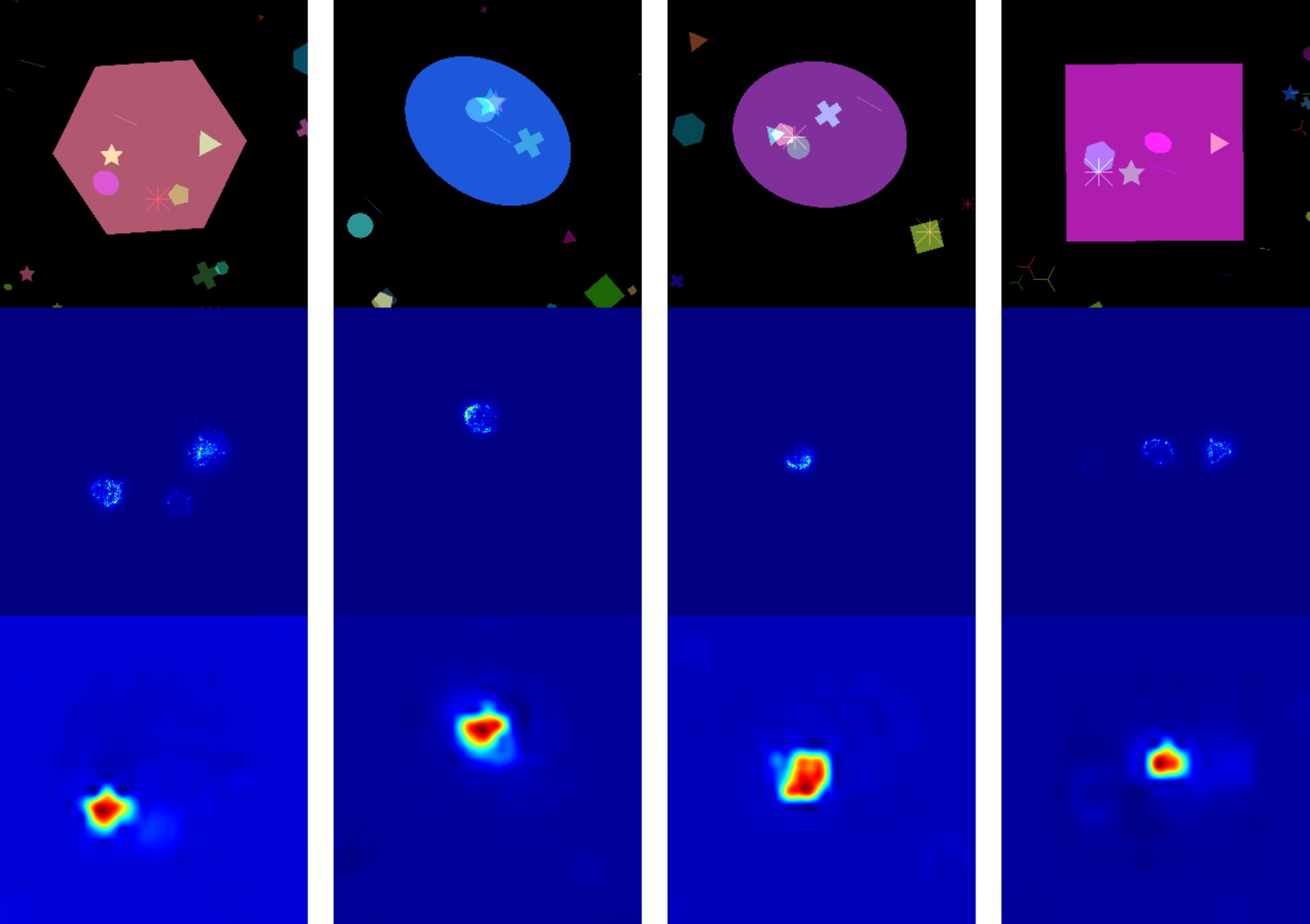}}\\
\subfloat[Random images along with their generated CLMs for concepts \textit{hexagon} (first two columns) and \textit{star} (second two columns). Middle row shows g-CLM (SG-SQ) and lower row shows p-CLM.]{\label{fig:ToyEvalB} \includegraphics[width=.8\textwidth]{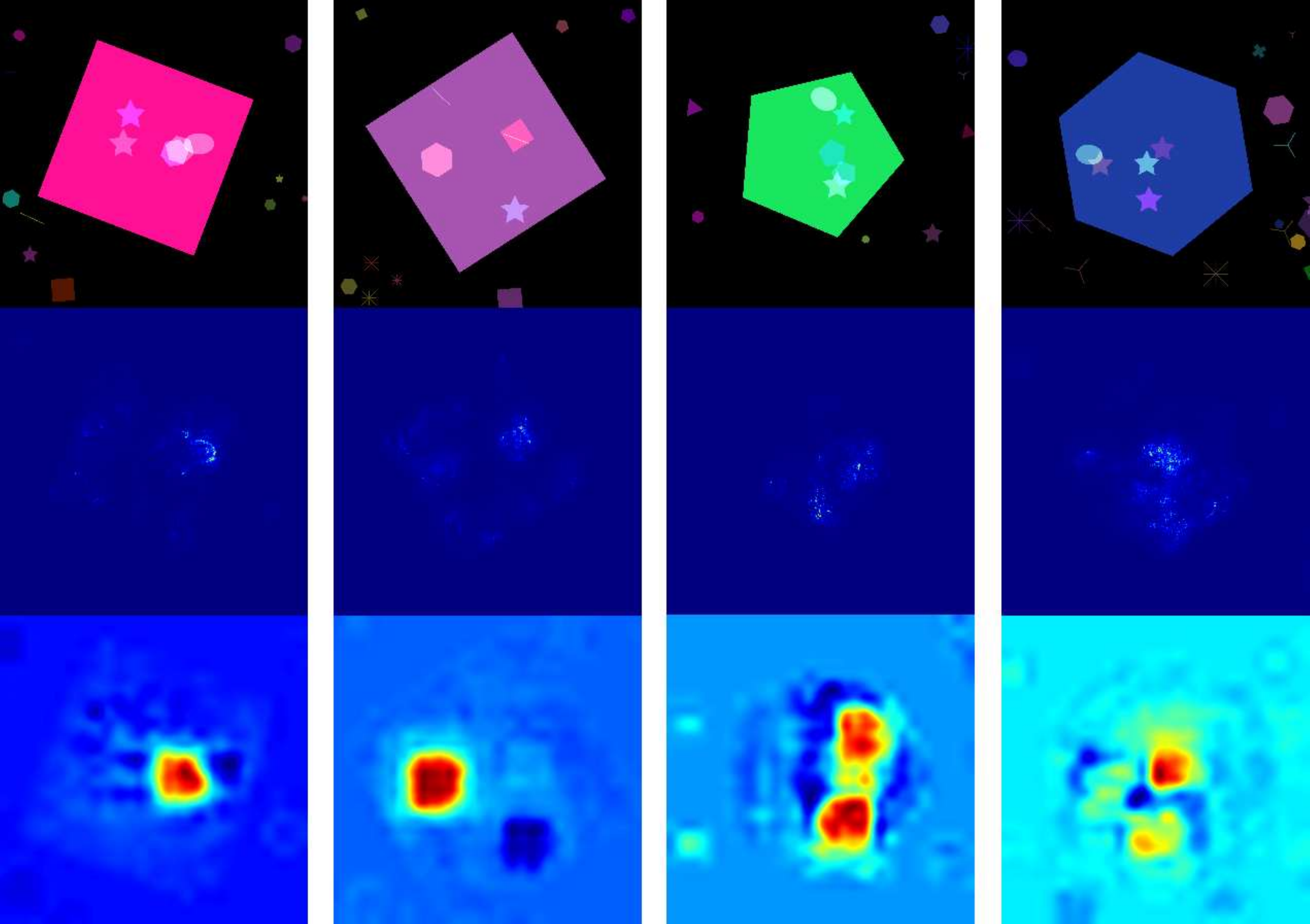}}
\caption{Images from proposed SCDB dataset shown in first rows along with corresponding concept localization maps from \textit{SE-ResNeXt-50} on layer \textit{pool5}.} \label{fig:ToyEval}
\end{figure}

\subsection{Evaluation on Synthetic SCDB Dataset}
\begin{table}[!b]
\centering
\caption{Test accuracies on SCDB dataset and average concept accuracies on \textit{pool5} layer of each architecture.}
\label{tab:ToyAccuracy}
\begin{tabular}{l|c|c|c}
                                  & ~VGG16~ & ~ResNet50~ & ~SE-ResNeXt-50~ \\ \hline
Image Classification Accuracy {[}\%{]}            & 97.5  & 93.5     & 95.6          \\
Concept Classification Accuracy {[}\%{]} & 85.7  & 81.1     & 72.8         
\end{tabular}
\end{table}

The resulting accuracies of models trained on SCDB are shown in Table~\ref{tab:ToyAccuracy}. Surprisingly, the simplest and shallowest architecture achieved the highest test accuracy. 
%[This might be the case because HPT was done on that architecture. PROBABLY NO: SEResNeXt with Adam achieved only as well 96.5]. 
However, the average concept classification accuracies on the architectures' last pooling layers (\textit{pool5}) indicate that complex architectures posses more informed representations of concepts. Fig.~\ref{fig:ToyEval} shows some examples of SCDB along with generated CLMs. Rows two and three correspond to g-CLM (SG-SQ) and p-CLM, respectively. The examples presented in this figure reveal that g-CLMs can be used to localize concepts in many cases. However, it appears that the method often highlights additional \textit{biomarkers} that do not correspond to the investigated concept. For some concepts, localization failed for almost all examples. Furthermore, the generated maps appear to be sparse and distributed, which is typical for methods based on input gradients. The heatmaps obtained from p-CLM are extremely meaningful and descriptive, as shown in lower rows of Figures~\ref{fig:ToyEvalA} and~\ref{fig:ToyEvalB}. The granularity of these heatmaps is restricted by the computational cost (through chosen patch-size and stride) as well as the average concept size on the image. It is evident that the method is able to separate the contributions of specific image regions to the prediction of a certain concept. This even holds true if shapes are overlapping.

\subsubsection{Quantitative Evaluation:} 
\begin{figure}[!t]
\centering
\includegraphics[width=1\textwidth]{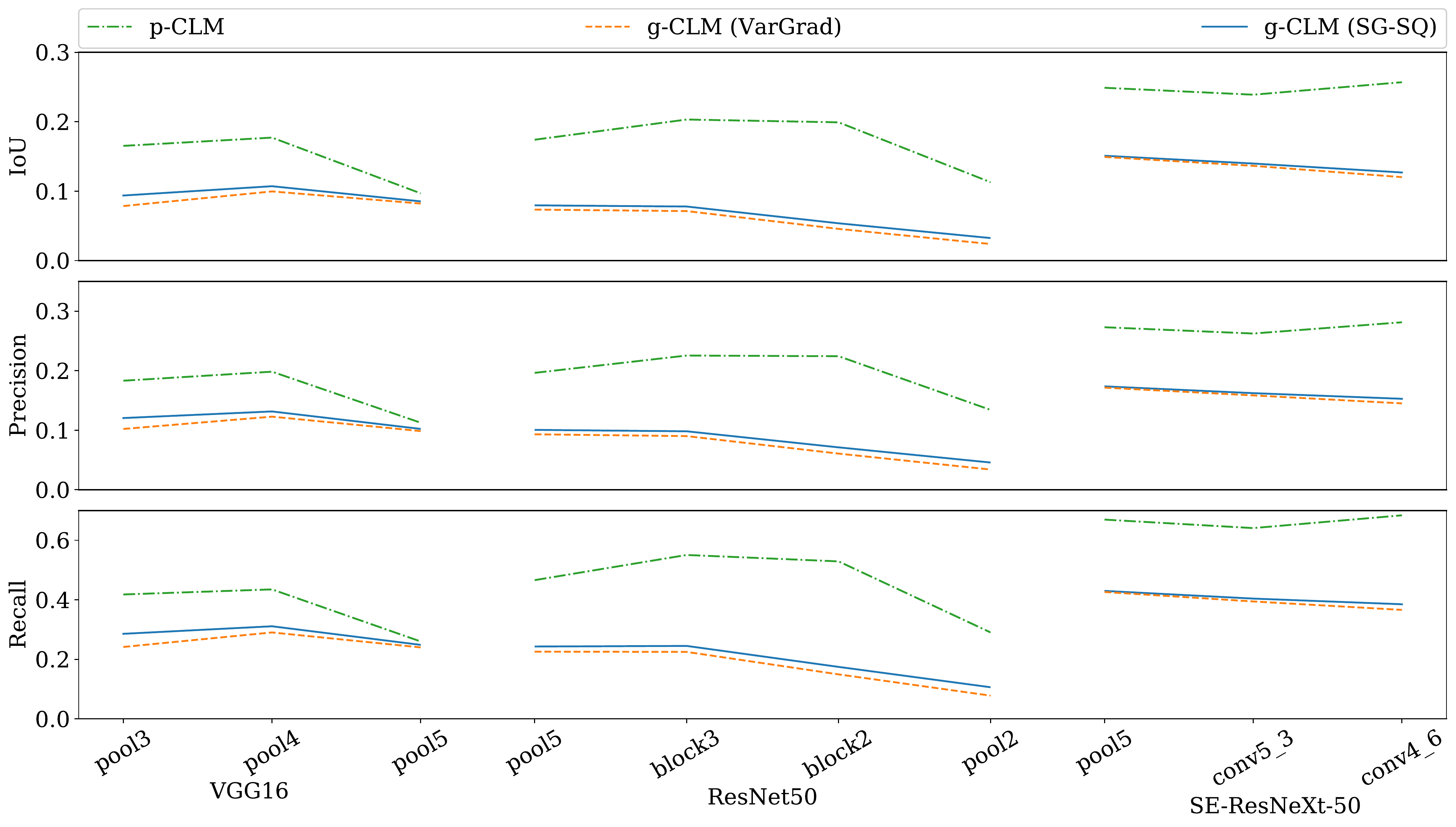}
\caption{Average IoU, precision and recall over all 10 concepts for predicted CLMs applied to three network architectures.}
\label{fig:ToyQuantitative}
\end{figure}

%The qualitative results reveal the feasibility of the proposed method. 
To quantify CLMs performance, we compute average IoU, precision and recall between predicted CLMs and their respective ground truth masks for all images in the validation set of SCDB dataset. Therefore, the predicted CLMs are binarized using a per-map threshold from the 98\% percentile. The metrics are computed for all images with a positive concept ground truth which means that images with incorrect concept prediction are included as well. 
%Here, only the perturbation-based CLM is considered, as the sparsity of gradient-based CLMs and remaining noise lead to poor quality of binarization maps. 
Average results over all 10 concepts for all networks and variants are presented in Fig.~\ref{fig:ToyQuantitative}. Concept localization performance of all methods increased with model complexity. This suggests that concept representations are most accurate in \textit{SE-ResNeXt-50}. Results also clearly show that both variants of g-CLM are outperformed by p-CLM over all networks. p-CLM achieved best average localization recall of 68\% over all 10 concepts, followed by g-CLM (SG-SQ) with 38\% and g-CLM (VarGrad) with 36\%. Most concepts relevant to the classification achieved recalls over 80\% with p-CLM. The best IoU of 26\% is also scored by p-CLM. It needs to be noted that IoU is an imperfect measure considering the sparsity and gradient-based CLMs and the granularity of p-CLM.

%IoU scores are mostly low with \textit{starmarker} achieving the best value of \( 0.38 \). This has several reasons. First, the patch size for occlusion was chosen to be \( 30 \times 30 \) pixels to avoid excessive computation time, resulting in relatively coarse CLMs. Secondly, the IoU might be tuned by further tuning the binarization threshold. However, a better indicator for the viability of the method is the recall, as it describes the portion of concept pixels that are correctly localized. 

%For many concepts, recalls over 80\% are achieved. It is striking, that \textit{SE-ResNeXt-50} and \textit{ResNet50} show better overall concept localization. This includes concepts like \textit{cross} and \textit{line} that are uninformative for the target task. In addition, \textit{SE-ResNeXt-50} is the only architecture that shows constant localization performance over the last three layers. 

Both qualitative and quantitative analyses suggest that the performance of CLM and thus the representation of concepts is improved with the complexity of the model architecture. 
%This finding is contrary to recent claims by Hu et al.~\cite{hu2020architecture}. They concluded that simpler architectures allow for easier disentanglement and are therefore more interpretable, comparing VGG16 to ResNet and DenseNet~\cite{huang2017densely}.

%%%%%%%%%%
% CelebA %
%%%%%%%%%%

\subsection{Evaluation on CelebA Dataset}
\begin{figure}[h!]
\centering
\subfloat[Images with correct prediction of concept \textit{lipstick}. First row shows the original image with the heatmap overlay of p-CLM, second row the g-CLM (SG-SQ) and last row p-CLM.]{\label{fig:CelebALipstick} \includegraphics[width=.95\textwidth]{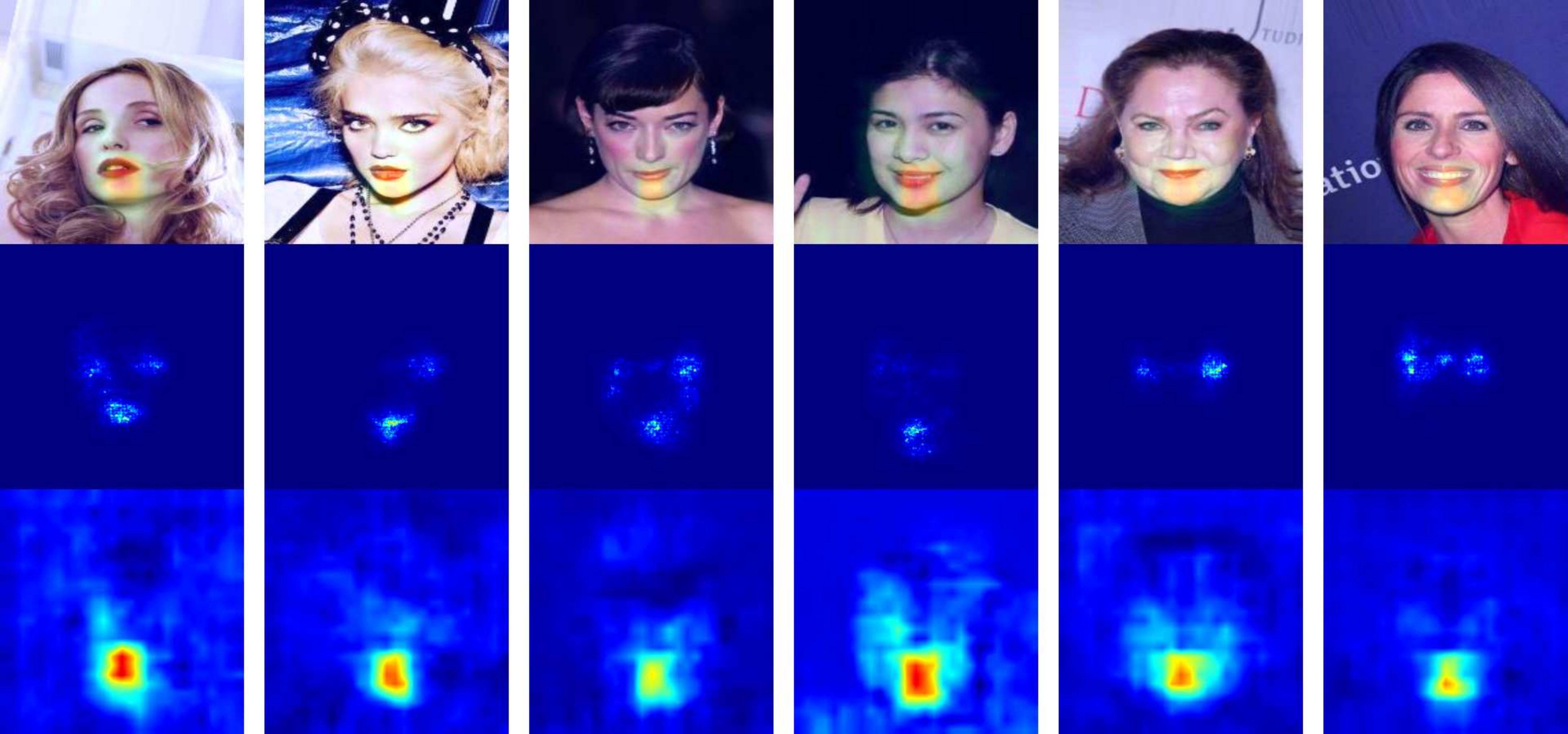}}\\

\subfloat[Images with positive predictions of concept \textit{bald}. First row shows the original image with the heatmap overlay of p-CLM, second row the g-CLM (VarGrad) and last row p-CLM.]{\label{fig:CelebABaldness} \includegraphics[width=.95\textwidth]{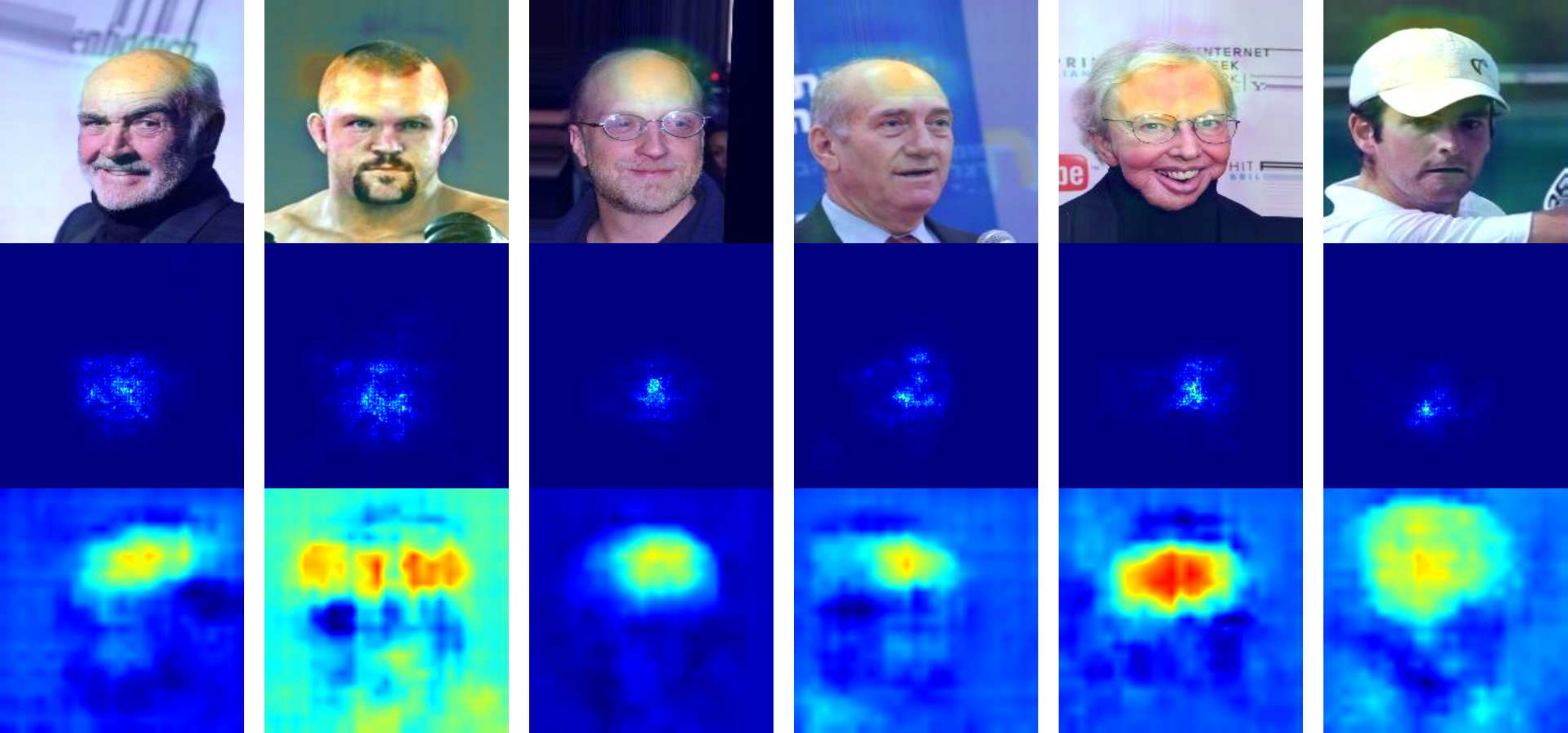}}\\

\subfloat[Images with positive predictions of concept \textit{hat} along with with their CLMs. Left column shows the original image with overlay of p-CLM, middle column shows g-CLM (SG-SQ) and right column shows p-CLM.]{\label{fig:CelebAHats} \includegraphics[width=.95\textwidth]{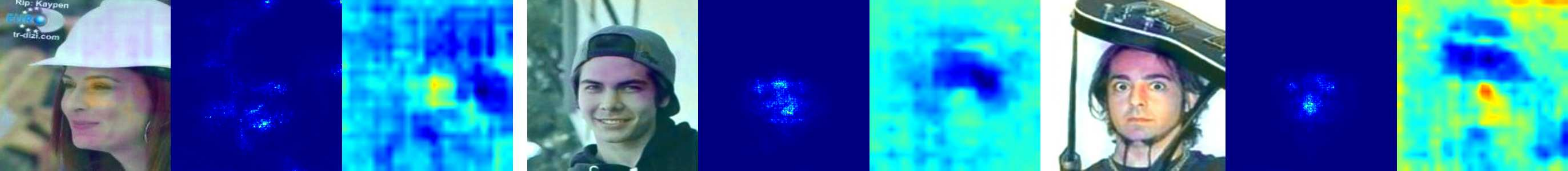}}

\caption{Examples for CLMs generated from \textit{SE-ResNeXt-50} trained on binary classification of gender with CelebA dataset.}
\label{fig:CelebAQualitative}
\end{figure}

Learning from our experiments on SCDB, we trained only \textit{SE-ResNeXt-50} model on the binary gender classification task in CelebA. The resulting network achieved 98.6\% accuracy on the test split. Concepts that achieved highest accuracies are often strongly related to single classes like facial hair (e.g. \textit{goatee}, \textit{mustache}, \textit{beard} and \textit{sideburns}) or makeup (e.g. \textit{heavy makeup}, \textit{rosy cheeks} and \textit{lipstick}). Fig.~\ref{fig:CelebAQualitative} shows images with their corresponding CLMs generated with our method. Due to the absence of ground truth segmentation masks in this dataset, no quantitative evaluation can be made. However, qualitative evaluation of some particularly interesting concepts is discussed below.

\subsubsection{Lipstick:} Fig.~\ref{fig:CelebALipstick} shows examples of CLMs for the \textit{lipstick} concept. Although it is quite likely that the network learnt a more abstract notion of female and male lips for the classification, the robust localization indicates that the network indeed encodes a lip related concept in the learnt CAV direction. It is striking that g-CLM often fails, highlighting the cheeks as well.

\subsubsection{Facial Hair:} All concepts related to facial hair achieved concept accuracies exceeding 80\%. However, inspecting the generated concept localization maps reveals that the CAVs do not properly correspond with the nuances in concept definitions. 
%Figure~\ref{fig:CelebAFacialHair} shows CLMs of concepts \textit{goatee}, \textit{mustache} and \textit{sideburns}. 
The localization maps reveal that the concept \textit{sideburns} never actually locates sideburns but beards in general. For the \textit{goatee} and \textit{mustache} it can be observed that a distinction between both is rarely made. It is thus very likely that the network learned a general representation of facial hair instead of different styles, as it would not aid solving the target task of classifying males versus females.

\subsubsection{Bald:} The \textit{bald} concept produces almost perfect p-CLMs focusing on the forehead and bald areas. It perfectly demonstrates how the network learnt an intermediate-level feature from raw input that is strongly correlated to a target class. 
%However, this also reveals a bias towards male class when detecting too much forehead. 
An intriguing finding is that often times, hats are confused with baldness as seen in the last column of figure~\ref{fig:CelebABaldness}. However, g-CLM consistently failed to locate this concept.

\subsubsection{Hats:} Despite of being sometimes mistaken for baldness, the localization maps for \textit{hats} in figure~\ref{fig:CelebAHats} prove that the network struggled to learn correct representation of a hat.

\section{Conclusion}
\label{sec:Conclusion}
%Reliable localization of human-understandable concepts learnt by DNNs for prediction is a valuable tool aiding explanation of DNNs and their use as decision support systems. 
We introduced the novel direction of concept localization for explanation of AI-based DSS and proposed a robust perturbation-based concept localization method (p-CLM) that has been evaluated on a synthetically generated dataset as well as a publicly available dataset of natural face images. p-CLM considerably outperformed two gradient-based variants (g-CLM) in qualitative and quantitative evaluation. Our initial results are promising and encourage further refinement of this approach. The computational efficiency and quality of heatmaps can be greatly improved by utilizing optimization-based perturbation methods like~\cite{fong2019understanding} and~\cite{fong2017interpretable}. Not only will they reduce the number of network propagations by optimizing the prediction score, but also the flexible shape of masks would be beneficial for the quality of CLMs. Perturbation-based methods always introduce some distribution shift which might distort predicted outcomes. However, more sophisticated methods like image impainting could minimize distribution shifts through perturbation. The method of CLM is another step towards explainable AI that could help break the barriers in the way of practical utilization of AI-based solutions. Our SCDB dataset is also publicly available for research community to advance XAI using concept based interpretation.

%A basic requirement of many concept-based explanation methods including the proposed CLM is the availability of concept annotations. Therefore, it is worth striving for concept annotated subsets of existing datasets to allow for concept explanation of their classifiers. Another feasible alternative would be the additional annotation of segmentation datasets for specific classification tasks.

\bibliographystyle{splncs04}
\bibliography{references}

\end{document}